\pdfoutput=1

\documentclass[11pt]{article}

\usepackage{acl}

\usepackage{times}
\usepackage{latexsym}

\usepackage[T1]{fontenc}

\usepackage[utf8]{inputenc}

\usepackage{microtype}

\usepackage{graphicx}
\usepackage{float}
\usepackage{amsmath}

\usepackage{caption}
\usepackage{subcaption}
\usepackage[T1]{fontenc}
\usepackage[utf8]{inputenc}
\usepackage{microtype}
\usepackage{amssymb}
\usepackage{array,multirow}
\usepackage{afterpage}

\newcommand{\data}[1]{\texttt{#1}}

\title{Segmenting Numerical Substitution Ciphers}
\author{Nada Aldarrab \and Jonathan May  \\
University of Southern California \\
Information Sciences Institute \\
\texttt{\{aldarrab,jonmay\}@isi.edu}}

\begin{document}
\maketitle
\begin{abstract}

Deciphering historical substitution ciphers is a challenging problem. Example problems that have been previously studied include detecting cipher type, detecting plaintext language, and acquiring the substitution key for segmented ciphers. However, attacking unsegmented, space-free ciphers is still a challenging task. Segmentation (i.e. finding substitution units) is the first step towards cracking those ciphers. In this work, we propose the first automatic methods to segment those ciphers using Byte Pair Encoding (BPE) and unigram language models. Our methods achieve an average segmentation error of 2\% on 100 randomly-generated monoalphabetic ciphers and 27\% on 3 real homophonic ciphers. We also propose a method for solving non-deterministic ciphers with existing keys using a lattice and a pretrained language model. Our method leads to the full solution of the IA cipher; a real historical cipher that has not been fully solved until this work.

\end{abstract}

\section{Introduction}

The contents of thousands of historical documents are still unknown to the contemporary age, even though they are encrypted using classical methods. Example documents include books from secret societies, diplomatic correspondences, and pharmacological books. Previous work has been done on collecting historical ciphers from libraries and archives and making them available for researchers \citep{pettersson-megyesi-2019-matching, megyesi-2020}. However, decipherment of classical ciphers is an essential step to reveal the contents of those historical documents. 

\begin{figure*}	
    \centering
	\includegraphics[scale=0.21]{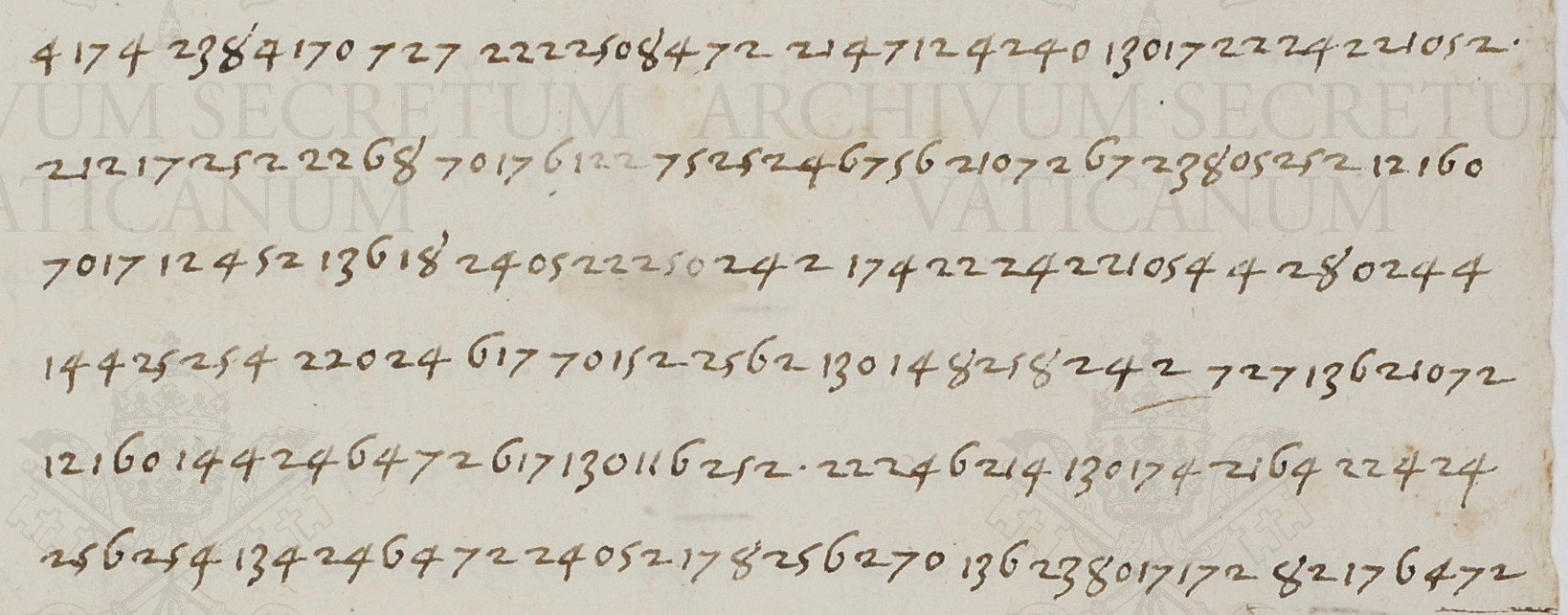}
	\caption[The IA cipher (16th century]{The IA cipher (16th century).\footnotemark} 	
	\label{fig:ia}		
\end{figure*}

Deciphering historical substitution ciphers has attracted attention from the natural language processing community. Example work includes \cite{ravi-2008, corlett-2010, nuhn-2013, nuhn-2014, hauer-2014, aldarrab-2017, kambhatla-2018, aldarrab-may-2021-sequence}. However, these methods all assume that cipher elements are clearly segmented (i.e., that token boundaries are well established). Many historical documents, however, are enciphered as continuous sequences of digits that hide token boundaries \citep{lasry-20}. An example cipher (the IA cipher) is shown in Figure~\ref{fig:ia} \cite{megyesi-2020}. Solving those ciphers is very challenging since it is not possible to directly search for the key without finding substitution units.  We use the term \textbf{numerical ciphers} to refer to unsegmented, space-free substitution ciphers that use a numerical symbol set.\footnote{The proposed methods can of course be applied to any unsegmented, space-free substitution cipher, regardless of the chosen symbol set.} 

In this work, we address the problem of segmenting numerical ciphers. The contributions of our work are:
\begin{itemize}
     \item We propose the first model to segment non-deterministic numerical ciphers \textbf{with existing keys} using a segmentation lattice and a pretrained language model. Our method unveils the content of the IA cipher; a letter from the 16th century that has not been revealed until this work.
    \item We propose novel unsupervised methods to segment numerical ciphers \textbf{with no existing keys} using Byte Pair Encoding (BPE) \citep{gage-94} and unigram language models \cite{kudo-2018-subword}. 
    \item We conduct extensive testing of our methods on different cipher types. We report results on synthetic and real historical ciphers and show how performance varies with cipher type and length. Our methods achieve an average segmentation error of 2\% on 100 randomly-generated monoalphabetic ciphers and 27\% on 3 real homophonic ciphers.
\end{itemize}

\afterpage {\footnotetext{\url{https://de-crypt.org/decrypt-web/RecordsView/189?showdetail=}}}

\section{Problem Definition}

A substitution cipher is a cipher that is created by substituting each plaintext character with another character according to a substitution table called the \textbf{key}. We define major terms in the following subsections.

\subsection{Substitution types}

In this paper, we focus on two types of substitution ciphers: Monoalphabetic and homophonic ciphers. Monoalphabetic ciphers are created by replacing each plaintext character with a unique substitute using a 1$\to$1 substitution key. Homophonic ciphers are created by replacing each plaintext character with one of multiple possible substitutes using a 1$\to$M substitution key.

\begin{figure*}	
    \centering
	\includegraphics[width=339px]{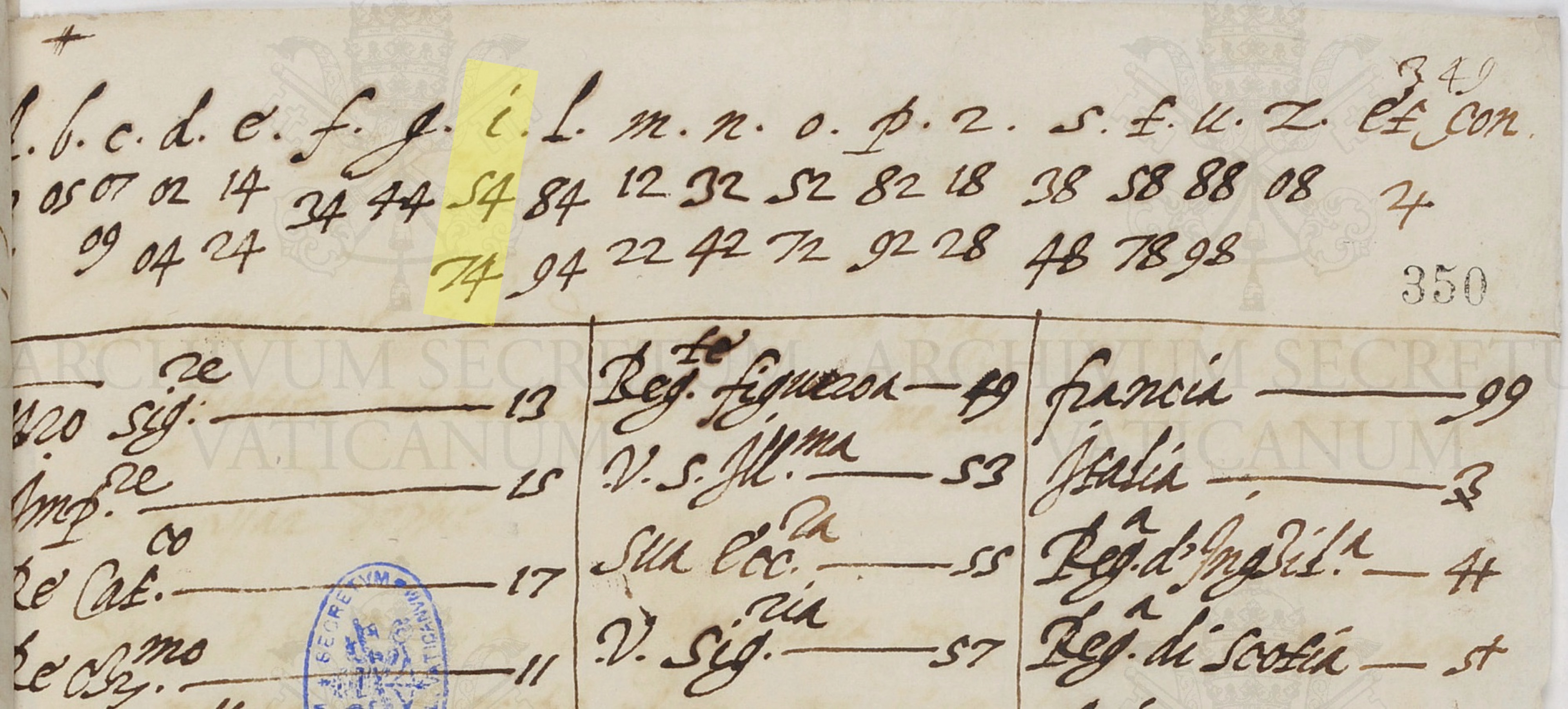}
	\caption[An example homophonic key from the Vatican Secret Archives (16th century)]{An example homophonic key from the Vatican Secret Archives (16th century).\footnotemark{} The added highlight in the top section shows that, e.g. \data{i} can be substituted with \data{54} or \data{74}. The bottom section contains nomenclature elements.} 	
	\label{fig:nom-key}		
\end{figure*}

For example, the key shown in Figure \ref{fig:nom-key} contains a homophonic substitution table (the top part). As shown in the figure, each plaintext character (e.g.~\data{i}) can be substituted with one of multiple characters (e.g.~\data{54} or \data{74}). It is common to encipher vowels with more than one character, which makes homophonic ciphers harder to crack. 

\afterpage{\afterpage {\footnotetext{\url{https://de-crypt.org/decrypt-web/RecordsView/206?showdetail=}}}}

\subsection{Cipher elements}

Cipher elements are substitution units that correspond to plaintext elements according to a cipher key. There are three main types of cipher elements in historical ciphers:

\begin{itemize}
    \item \textbf{Regular elements:} These elements usually encode letters, common syllables, or prepositions. In the example key shown in Figure \ref{fig:nom-key}, the top part defines regular cipher elements.
    \item \textbf{Nomenclature elements:} This refers to elements in a key that represent whole words (often proper names). In Figure \ref{fig:nom-key}, the second part defines nomenclature elements.
    \item \textbf{Nulls:} These are cipher elements that do \textbf{not} correspond to any plaintext word or character. Nulls are usually used in ciphers to confuse cryptanalysts. Sometimes, nulls are used for a purpose. For example, they could be used to mark the beginning of nomenclature elements in numerical ciphers.
\end{itemize}

\subsection{Fixed and variable-length ciphers}

Numerical ciphers can be classified as fixed or variable-length ciphers. In fixed-length ciphers, regular elements have the same length (i.e. the same number of digits). However, in variable-length ciphers, regular elements can be of different lengths. For example, the letter \data{a} might be enciphered as \data{1}, \data{12}, or \data{121}.

\subsection{Ciphertext segmentation}
\label{sec:cipher-segmentation}

Numerical ciphers impose a special challenge, in that they hide cipher element boundaries. For example, in the numerical cipher shown in Figure \ref{fig:ia}, it is unclear which digits represent substitution units. Identifying substitution units, which we call \textbf{segmentation}, is a challenging task that is necessary to solve these ciphers.
Another challenge in solving numerical ciphers is that the segmentation could be non-deterministic. For example, a cipher can have these substitutions in its key:

\begin{table}[H]
\centering
\begin{tabular}{c|c}
 \textbf{Cipher} & \textbf{Plain} \\
\hline
\data{2} & \data{a}  \\
\data{22} & \data{n} \\
\data{8} & \data{d} \\
\hline
\end{tabular}
\end{table}

\noindent which means that the ciphertext \data{2228} can be segmented as:

\begin{table}[H]
\centering
\begin{tabular}{c|c}
 \textbf{Cipher Segmentation} & \textbf{Plain} \\
\hline
\data{2 | 2 | 2 | 8} & \data{a a a d}  \\
\data{2 | 22 | 8} & \data{a n d} \\
\data{22 | 2 | 8} & \data{n a d} \\
\hline
\end{tabular}
\end{table}

Such ciphers are called \textbf{non-deterministic ciphers.} \textbf{Deterministic ciphers}, on the other hand, only have one possible segmentation according to their keys.

In this paper, we focus on the problem of segmenting numerical ciphers. We look at two cases for numerical cipher segmentation depending on whether or not a key exists for the cipher in hand. The following sections describe our proposed methods for each case.

\section{Segmenting Non-Deterministic Ciphers with an Existing Key}
\label{sec:seg-existing-key}

We start with the first case: Suppose we have a cipher and a key, but the cipher is non-deterministic. This case can arise in practice when the key of the cipher is found while combing through historical archives, for example. Alternatively, the key could have been found by a cryptanalyst by solving a part of the cipher. Although the cipher key exists in these scenarios, the non-deterministic segmentation makes it impossible to directly apply the key to recover the plaintext (Recall the ambiguous segmentation example of the word \data{and} from Section~\ref{sec:cipher-segmentation}). In this case, it is very challenging to manually recover the whole plaintext, especially when the cipher is very long.

\subsection{Lattice segmentation}
\label{sec:lattice-seg}

We take as an example the IA cipher (Figure~\ref{fig:ia}), which we retrieved from the DECRYPT database \citep{megyesi-2020}. The first few lines of this 16th-century cipher were deciphered in 2019. However, since the cipher is non-deterministic, the remaining ciphertext (more than 200 lines) has not yet been deciphered. 

This is a real use case for our proposed method; a real historical cipher with an existing key but with a non-deterministic segmentation. For example, consider this part of the IA cipher key:

\begin{table}[H]
\centering
\begin{tabular}{c|c||c|c}
\hline
\textbf{Cipher} & \textbf{Plain} & \textbf{Cipher} & \textbf{Plain} \\
\hline
\data{0} & \data{e} & \data{22} & \data{p} \\
\data{2} & \data{o} & \data{24} & \data{r} \\
\data{4} & \data{a} & \data{25} & \data{t} \\
\data{5} & \data{s} & & \\
\hline
    \end{tabular}
\end{table}

\begin{figure}
	\centering
	\begin{tabular}{llllllll}	

\multicolumn{8}{c}{\includegraphics[scale=0.4]{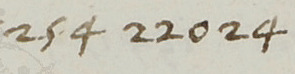}} \\

\data{2} & \data{5} & \data{4} & \data{2} & \data{2} & \data{0} & \data{2} & \data{4}  \\
\multicolumn{2}{l}{\data{25}} & \data{4} & \data{2} & \data{2} & \data{0} & \data{2} & \data{4}  \\
\data{2} & \data{5} & \data{4} & \multicolumn{2}{l}{\data{22}} & \data{0} & \data{2} & \data{4}  \\
\data{2} & \data{5} & \data{4} & \data{2} & \data{2} & \data{0} & \multicolumn{2}{l}{\data{24}}  \\
\multicolumn{2}{l}{\data{25}} & \data{4} & \multicolumn{2}{l}{\data{22}} & \data{0} & \data{2} & \data{4}  \\
\multicolumn{2}{l}{\data{25}} & \data{4} & \data{2} & \data{2} & \data{0} & \multicolumn{2}{l}{\data{24}}   \\
\data{2} & \data{5} & \data{4} & \multicolumn{2}{l}{\data{22}} & \data{0} & \multicolumn{2}{l}{\data{24}} \\
\multicolumn{2}{l}{\data{25}} & \data{4} & \multicolumn{2}{l}{\data{22}} & \data{0} & \multicolumn{2}{l}{\data{24}} \\
	\end{tabular}
	\caption[Example segmentation ambiguity for the IA cipher]{Example segmentation ambiguity for the IA cipher. A short 8-digit segment produces 8 possible segmentations according to the key. The number of candidate segmentations increases exponentially with respect to cipher length.}
	\label{fig:ia-ambiguity}

\end{figure}

Figure~\ref{fig:ia-ambiguity} shows a short 8-digit part of the IA cipher. As shown in the Figure, this part can be segmented in 8 possible ways according to the key. The number of candidate segmentations increases exponentially with respect to cipher length.

To solve this problem, we create a lattice to model all possible segmenations of the cipher using the existing key. Then we use a pretrained language model to choose the best possible segmentation (i.e. the segmentation that gives the most probable plaintext according to the language model). 

\begin{figure}	
    \centering
	\includegraphics[scale=0.1]{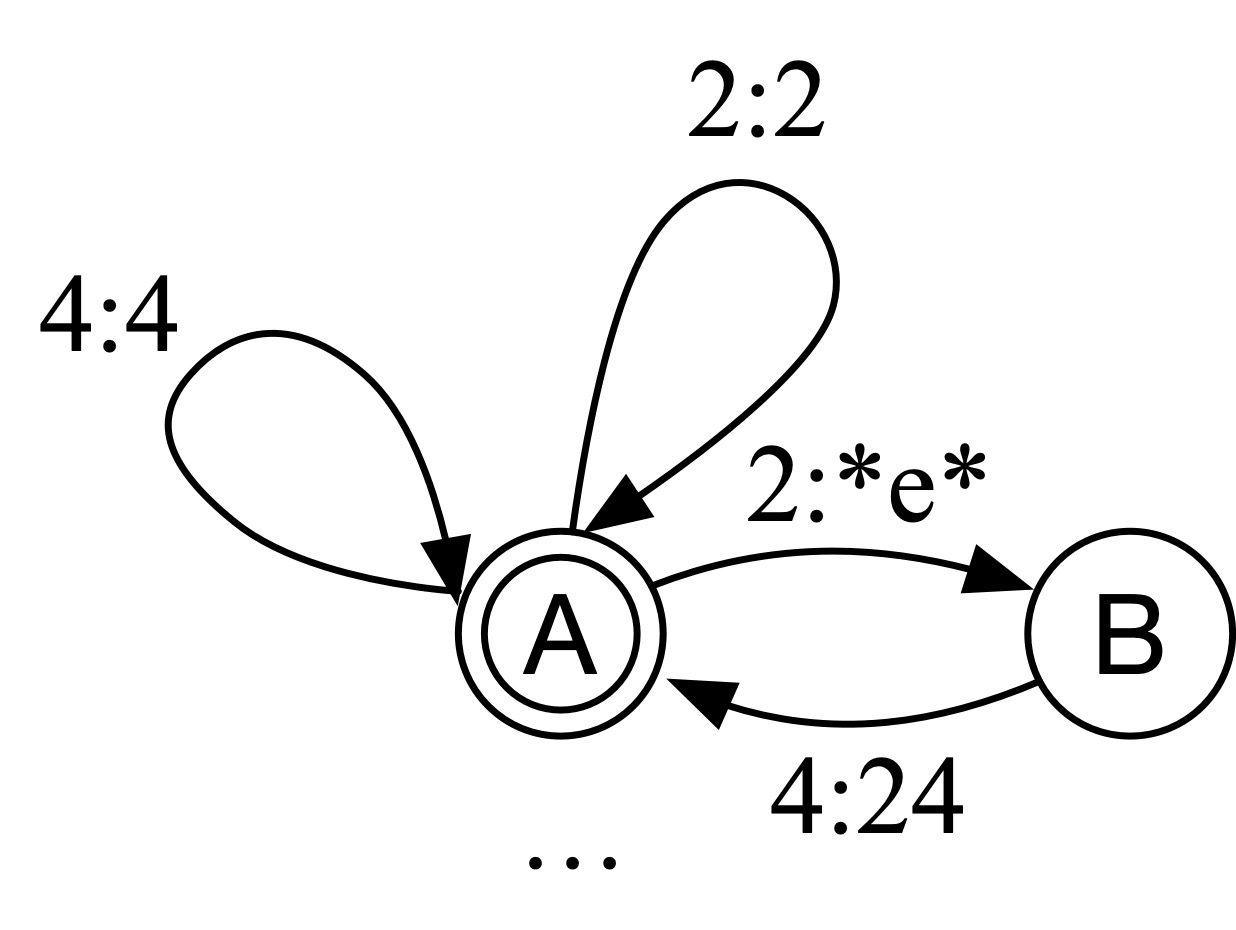}
	\caption{Part of the segmentation FST for the IA cipher. This part models the possiblity of merging the digits \data{2} and \data{4} to become \data{24} (corresponds to letter \data{r} in the key) vs. keeping them unmerged (letters \data{o} and \data{a} in the key). \data{*e*} is used to indicate the empty string.} 	
	\label{fig:seg-fst}		
\end{figure}

\begin{figure}	
    \centering
	\includegraphics[scale=0.1]{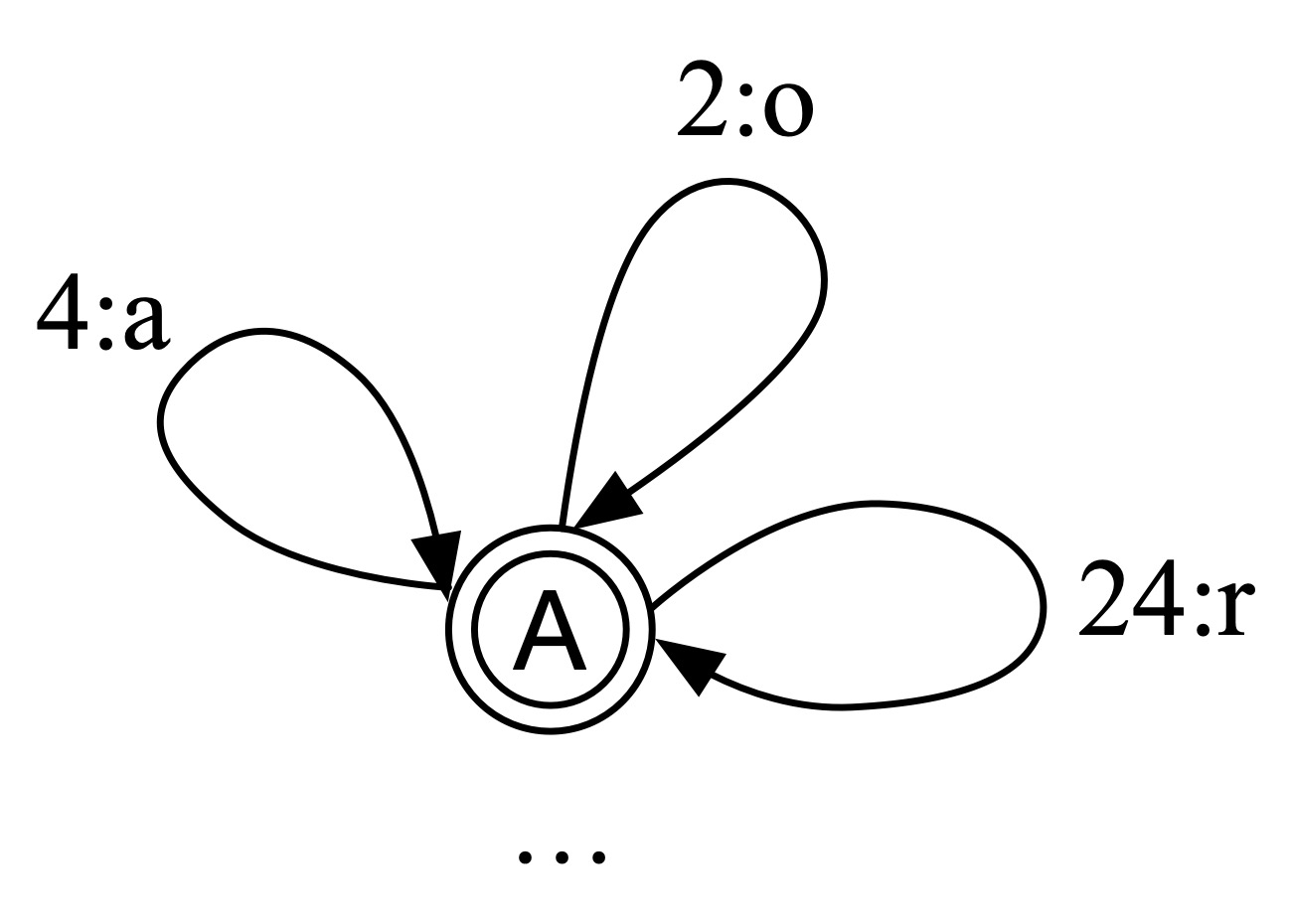}
	\caption{Part of the key FST for the IA cipher. This part models the substitutions: 2$\to$o, 4$\to$a, and 24$\to$r.} 	
	\label{fig:key-fst}		
\end{figure}

For the segmentation lattice, we create a Finite-State Transducer (FST) that models the possible merges of cipher symbols. Figure~\ref{fig:seg-fst} shows part of the FST. The shown transitions model the ambiguity of segmenting the digits \data{2} and \data{4}. According to the key, these two digits can be merged to become \data{24} (plaintext \data{r}) or stay unmerged (plaintext letters \data{o} and \data{a}, respectively). We create another FST to model the key (shown in Figure~\ref{fig:key-fst}).

We train a 5-gram character Italian language model on the historical data released by \citet{aldarrab-may-2021-sequence}. Composing the language model, key FST, and segmentation FST creates a lattice of all possible decipherments of the text. We use the Carmel finite-state toolkit to find the most probable plaintext according to the language model \cite{graehl-2010}. 

We follow \citet{aldarrab-may-2021-sequence} and use character-level Translation Edit Rate (TER) as our evaluation metric. TER is the character-level Levenshtein distance between system output and gold solution, divided by the number of characters in the gold solution. Verifying the resulting plaintext by a native Italian speaker, our method achieves a TER of 1.12\%, which means that our model's output is almost 99\% correct.\footnote{Prior to this experiment, we do not believe this plaintext had been known since 1536.}

\subsection{The IA cipher}
\label{sec:ia-cipher}

The resulting plaintext revealed a letter that the bishop of Senigallia sent to the Pope from Lisbon in 1536. 

The cipher is 11,026 characters. The key from DECRYPT included 21 cipher elements. However, decoding the rest of the letter revealed 4 more cipher elements (shown in Table~\ref{tab:ia-key}). We found out that the letter \data{e} can be enciphered as \data{0} or \data{.} in this cipher. Cipher element \data{19} seems to encode a nomenclature code. Cipher element \data{26} encodes the letter \data{x.} We also found that the symbol \data{$\therefore$} is used after the letters \data{du} to mean \data{ducati}, the currency used during that time.

\begin{table}
	\centering
	\begin{tabular}{c|c}	
\hline
\textbf{Cipher} & \textbf{Plain} \\ \hline
\data{0} or \data{.} &  \data{e}   \\
\data{19} &  (nomenclature element)  \\
\data{26} &  \data{x}    \\
$\therefore$ &  used after \data{du} to mean \data{ducati}  \\ \hline
	\end{tabular}
	\caption{Corrections/additions to the IA cipher key discovered by our approach.}
	\label{tab:ia-key}
\end{table}

We found out that there are human transcription errors in the transcription from DECRYPT. In total, we corrected 30 transcription errors in this cipher.

There also seem to be some errors in the original manuscript. Such errors can result from spelling mistakes or substitution mistakes during encipherment, for example. For those errors, we do not change the original ciphertext and consider the text as is.

\section{Segmenting Ciphers with no Existing Keys}

This is the second (and the more challenging) case; Segmenting a numerical cipher with no existing key. In this case, all we have is a sequence of digits (e.g. Figure~\ref{fig:homophonic-ciphers}). To solve the cipher, we first need to segment ciphertext before trying to find the substitution key. The rest of the paper focuses on this case.

\subsection{Baselines}
\label{sec:baselines}

We first try two baselines: 1-digit and 2-digit segmentation. We remove line breaks and consider the text as one long sequence of digits. In 1-digit segmentation, we split ciphertext into individual digits. In 2-digit segmentation, we split ciphertext into two-digit elements (except the last digit if the number of digits in the cipher is odd). The latter is a stronger baseline since we notice that most cipher elements in historical ciphers are two digits long.

\subsection{Byte Pair Encoding (BPE)}

Our first proposed method for cipher segmentation is Byte Pair Encoding (BPE). BPE is a simple compression algorithm that has been used for many natural language processing tasks \citep{gage-94,sennrich-etal-2016-neural}. In BPE, the most frequent pair of bytes is iteratively replaced with a single, unused byte to represent the replaced pair. The motivation behind using BPE for our problem is that the digits that belong to the same cipher element have high mutual information, so we would like them to be grouped together. 

\subsection{Unigram language model}

One downside of BPE is that it is a greedy algorithm that employs a deterministic symbol replacement strategy. BPE does not provide multiple possible segmentations with probabilities. As we notice from our experiments (Section~\ref{sec:mono-ciphers-sp}), the resulting BPE segmentation leaves many singleton digits unpaired.

To mitigate this problem, we use the subword segmentation algorithm proposed by \citet{kudo-2018-subword}, which is based on a unigram language model. This algorithm provides candidate segmentations with probabilities. The unigram language model assumes that each subword occurs independently. Thus, the probability of a subword sequence is the product of the subword probabilities. The probabilities are iteratively estimated using the Expectation Maximization (EM) algorithm. The most probable subword segmentation is then found by the Viterbi algorithm.

We evaluate the two baselines and our proposed methods on synthetic and real historical ciphers. The following sections describe our datasets, experiments, and results.

\section{Data}
\label{sec:data-seg}

To evaluate our methods on monoalphabetic numerical ciphers, we create synthetic ciphers from English Wikipedia.\footnote{Data will be released upon acceptance of the paper.} We notice that most historical ciphers in the DECRYPT collection are two pages long and contain about 2K characters, so we choose cipher length 2,048 for our experiments. We create 100 English ciphers using randomly generated keys. We use the numbers from 0 to 99 as possible cipher elements. This creates variable-length ciphers when single digits are chosen in the key. We report the average scores of the 100 ciphers for each experiment.

\begin{figure}	
\captionsetup[subfigure]{labelformat=brace}
\begin{subfigure}{0.5\textwidth}		
	\includegraphics[scale=0.1475]{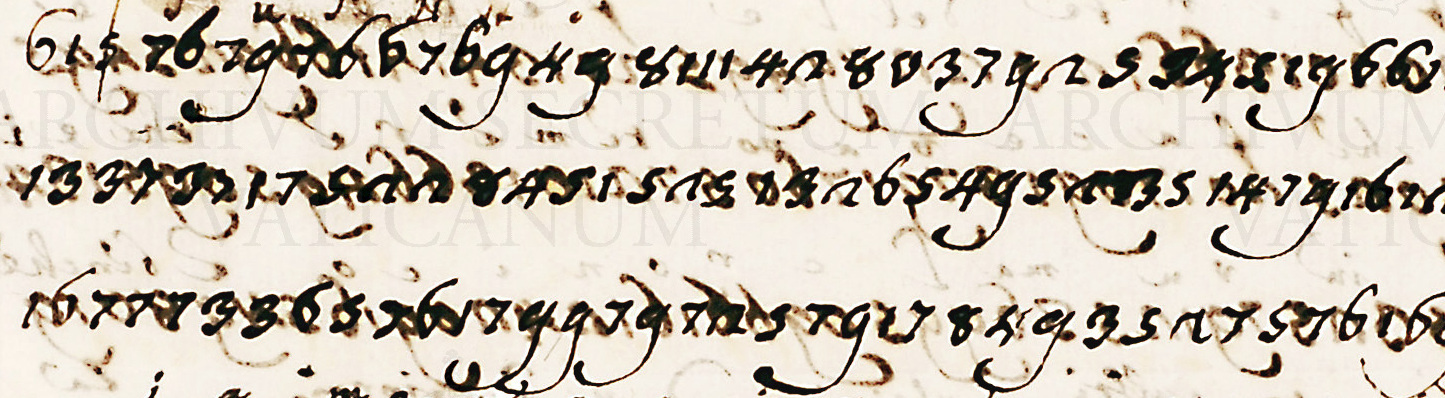}
	\caption{The S304 cipher (18th century).} 	
\end{subfigure}

\begin{subfigure}{0.5\textwidth}		
	\includegraphics[scale=0.1465]{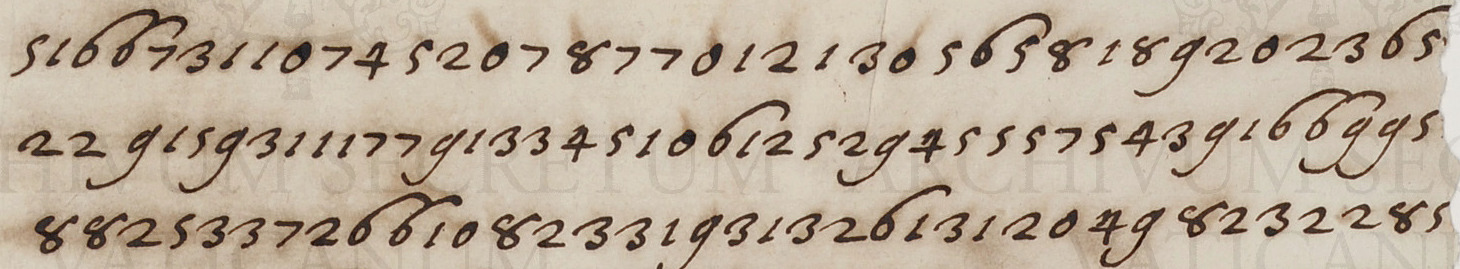}
	\caption{The C13 cipher (17th century).} 	
\end{subfigure}

\begin{subfigure}{0.5\textwidth}		
	\includegraphics[scale=0.1585]{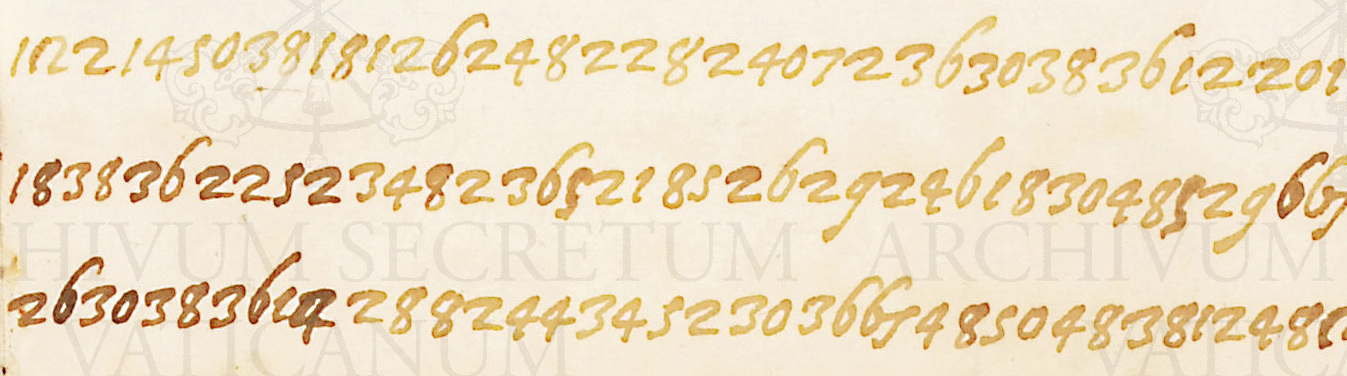}
	\caption{The F283 cipher (16th century).} 	
\end{subfigure}
	\caption[Three historical ciphers from the Vatican Secret Archives]{Three historical ciphers from the Vatican Secret Archives.\footnotemark} 	
	\label{fig:homophonic-ciphers}		
\end{figure}

For evaluation on real historical ciphers, we use 3 ciphers from the the Vatican Secret Archives, retrieved from the DECRYPT database \citep{megyesi-2020, lasry-20}. Table~\ref{tab:cipher-stats} shows cipher statistics. We use ciphers C13, S304, and F283 (Figure~\ref{fig:homophonic-ciphers}). For these ciphers, human transcriptions and gold segmentations are available on the DECRYPT database.

\begin{table*}
	\centering
	\begin{tabular}{lrrrrrr}	
\hline
Cipher & Length & Types & Tokens & 1-dig Tokens & 2-dig Tokens & 1+2-dig Tokens \\ \hline
S304 & 1,258  &  82 & 675 & 150 (22\%) & 496 (73\%) & 646 (96\%) \\
C13 & 1,879  &  97 & 917 & 0 (0\%) & 872 (95\%) & 872 (95\%) \\
F283 &  2,239  &  50  & 1,050 & 1 (0\%) & 979 (93\%) & 980 (93\%) \\
\hline
	\end{tabular}
	\caption[Statistics of the ciphers obtained from the DECRYPT database]{Statistics of the ciphers obtained from the DECRYPT database. Length is the number of digits in the cipher. Types and tokens are those of cipher elements, not individual cipher digits.}
	\label{tab:cipher-stats}
\end{table*}

\footnotetext{\url{https://de-crypt.org/decrypt-web/RecordsList}}

\section{Experimental Evaluation}

We carry out three types of experiments. First, we start with the simplest cipher type; monoalphabetic ciphers with spaces. The existence of spaces indicates word boundaries, which gives some clues on how to segment the ciphertext. Second, we remove spaces and try to segment the same monoalphabetic ciphers, which is expected to be a harder task. Third, we experiment with segmenting homophonic ciphers, which is the most challenging case discussed in this paper.

We apply our proposed segmentation methods on both synthetic and real historical ciphers. We also study the effect of cipher length on segmentation quality.

For evaluation, we use two metrics; F1 and Segmentation Error Rate (SegER). We use F1 to measure how good an algorithm is at finding cipher elements. For each algorithm, we compare the learned vocabulary with the gold vocabulary from the gold segmentation and report F1 scores.

We use SegER to evaluate the resulting segmented ciphertext after applying the segmentation algorithm. We define SegER as:

\begin{equation}
\mbox{SegER} = \frac{\mbox{\# of edits}}{\mbox{\# of reference segments}}
\end{equation}

\noindent where possible edits include the insertion, deletion, and substitution of single segments. 

We use the SentencePiece implementation of BPE and unigram language model \citep{kudo-richardson-2018-sentencepiece}. For homophonic ciphers, we set vocabulary size to the maximum number found by the unigram language model. For monoalphabetic ciphers, we set vocabulary size to 36 (26 maximum possible 2-digit elements for 1-1 substitutions + 10 singleton digits).

\subsection{Monoalphabetic ciphers with word spaces}
\label{sec:mono-ciphers-sp}

We first experiment with monoalphabetic ciphers with spaces. We test our methods on the 100 synthetic ciphers described in Section~\ref{sec:data-seg}. Table~\ref{tab:mono-sp-f1} (first column) shows F1 scores for all models. As expected, the 2-digit baseline is better than the 1-digit baseline, with an F1 score of 57\% as opposed to less than 14\% for the 1-digit baseline. BPE performs better than both 1-digit and 2-digit baselines, with an F1 score of about 65\%. 

As noted in Section~\ref{sec:baselines}, most cipher elements in historical ciphers are one or two digits long. In our random sample of three historical ciphers, about 95\% of cipher tokens are one and two-digit (as shown in Table~\ref{tab:cipher-stats}). Longer elements appear less often (less than 5\% of tokens in our test ciphers). Thus, we limit BPE piece length to a maximum of 2 digits. This improves the F1 score to about 81\%, with an improvement of about 25\% over default BPE. We call this model ``BPE 2'' in Table~\ref{tab:mono-sp-f1}.

We then apply the unigram language model of \citet{kudo-2018-subword}. We notice that the default unigram language model is subpar to default BPE. However, adding the 2-digit heuristic, we get an F1 score of 80\% using the unigram language model (Called ``Unigram LM 2'' in Table~\ref{tab:mono-sp-f1}), which is comparable to BPE~2.

\begin{table}
	\centering
	\begin{tabular}{l|rr}	
\hline
\textbf{Model} & \textbf{w/ spaces} & \textbf{w/o spaces} \\ \hline
1-dig baseline &  13.62  &  13.62  \\
2-dig baseline &  56.77  &  47.91 \\
BPE &  64.92  &  63.59  \\
BPE 2 &  \textbf{80.95}  &  79.59 \\ 
Unigram LM &  58.41  &  56.64 \\ 
Unigram LM 2 &  80.41  &  \textbf{80.71}  \\\hline
	\end{tabular}
	\caption[Average F1 \% ($\uparrow$) for segmenting 10 synthetic ciphers using different models]{Average F1 \% ($\uparrow$) for segmenting 100 synthetic ciphers using different models. In BPE 2 and Unigram LM 2, maximum piece length is set to 2.}
	\label{tab:mono-sp-f1}
\end{table}

We then apply the learned vocabularies to segment the ciphertexts. Table~\ref{tab:mono-sp-seger} (first column) shows SegER scores for all models. Default BPE does not perform better than the 2-digit baseline in terms of the resulting segmentation quality. However, BPE 2 gives a 53\% improvement over the 2-digit baseline, with a SegER of 10.95\%. The unigram language model with the 2-digit heuristic gives the best result of all models with a SegER of 2.45\%.

\begin{table}
	\centering
	\begin{tabular}{l|rr}	
\hline
\textbf{Model} & \textbf{w/ spaces} & \textbf{w/o spaces} \\ \hline
1-dig baseline &  181.05  &  181.05  \\
2-dig baseline &  23.20  & 49.89  \\
BPE &  34.32 &  36.74  \\
BPE 2 & 10.95  &  13.72 \\ 
Unigram LM &  40.28  &  41.61 \\ 
Unigram LM 2 &  \textbf{2.45}  &  \textbf{2.70}  \\\hline
	\end{tabular}
	\caption{Average SegER \% ($\downarrow$) for segmenting 100 synthetic ciphers using different models.}
	\label{tab:mono-sp-seger}
\end{table}

To better explain the motivation behind using the unigram language model, we show example BPE~2 errors in Figure~\ref{fig:example-bpe-errors}. These two examples come from the same cipher. For this cipher, BPE learned the right vocabulary elements of \data{17}, \data{71}, and \data{77}. However, since \data{77} is the most frequent of all, BPE always prefers to merge the two \data{7}s first. This early merge results in two unmerged single digits (\data{1} and \data{7} in the first example and two \data{1}s in the second example). This way, BPE~2 misses the correct merges of \data{17} and \data{71}. The unigram language model, on the other hand, looks at the overall score of segmentation candidates and chooses the most probable one according to unigram frequencies. In both examples, Unigram~LM~2 does a better job at segmenting ciphertext.

\begin{figure}
    \resizebox{\linewidth}{!}{
	\centering
	\begin{tabular}{l|lllllll}	
\hline
Gold & \texttt{86} & \texttt{1} & \multicolumn{2}{l}{\texttt{17}} & \texttt{77} & \texttt{65} & \texttt{39}  \\
BPE 2 & \texttt{86} & \texttt{1} & \texttt{1} & \texttt{\underline{77}}  & \texttt{7} & \texttt{65} & \texttt{39}  \\
Unigram LM 2 & \texttt{86} & \texttt{1} & \multicolumn{2}{l}{\texttt{17}} & \texttt{77} & \texttt{65} & \texttt{39}  \\ \hline
Gold & \texttt{65} & \texttt{17}  & \texttt{77} & \texttt{71} &&& \\
BPE 2 & \texttt{65} & \texttt{1} & \texttt{\underline{77}} & \texttt{\underline{77}} & \texttt{1} & \\
Unigram LM 2 & \texttt{65} & \texttt{17}  & \texttt{77} & \texttt{71} &&& \\ \hline
	\end{tabular}
}
	\caption[Example BPE segmentation errors]{Example BPE segmentation errors (incorrect merges underlined). In both examples, BPE chooses the wrong merges as it goes greedily from left to right. Unigram LM, on the other hand, looks at candidate segmentations and chooses the highest scoring candidate based on segmentation probabilities.}
	\label{fig:example-bpe-errors}
\end{figure}

\begin{figure}	
	\includegraphics[scale=0.45]{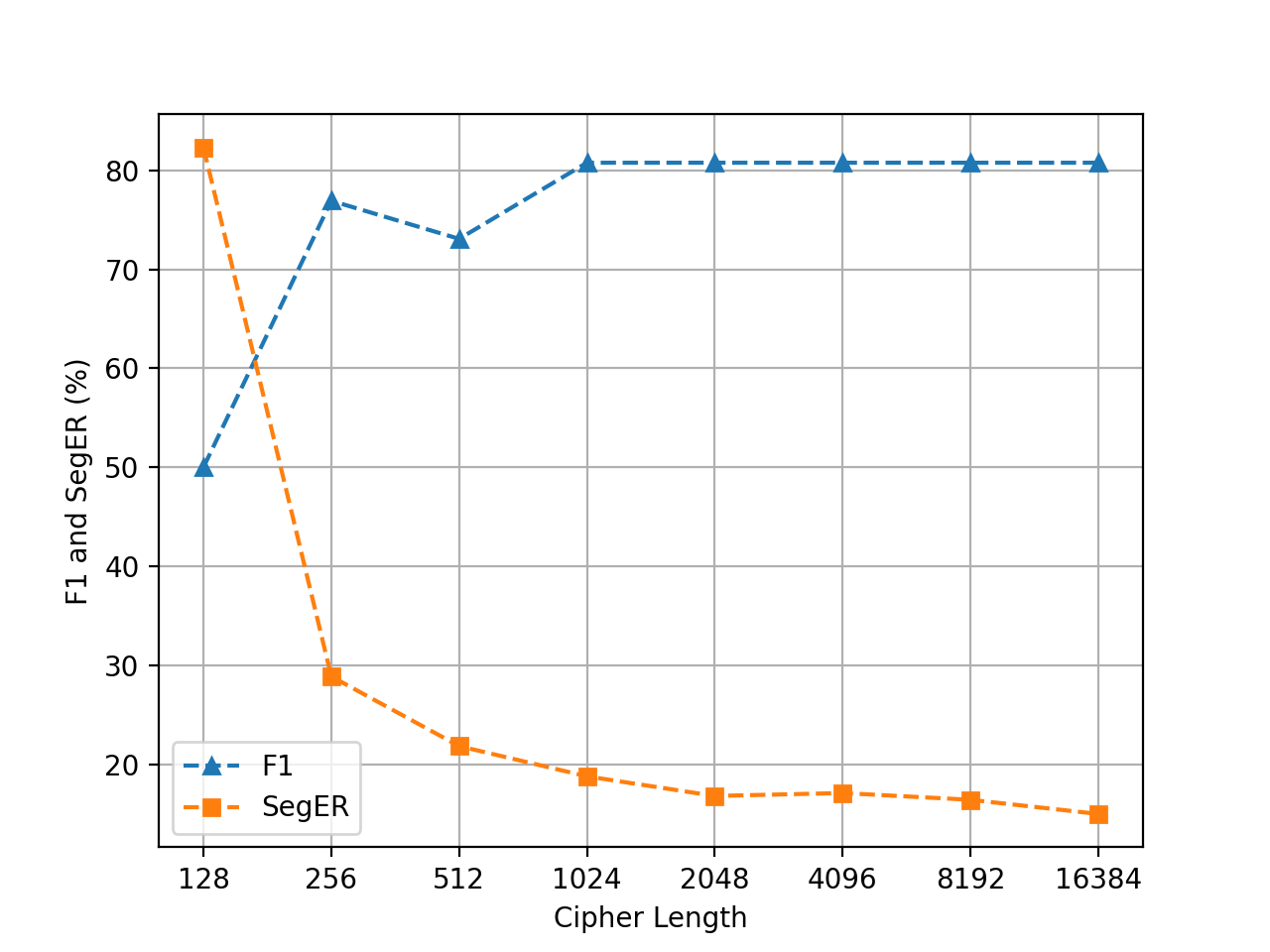}
	\caption{F1 \% ($\uparrow$) and SegER \% ($\downarrow$) for segmentation of different cipher lengths.}
	\label{fig:cipher-len}		
\end{figure}

To study the effect of cipher length on segmentation quality, we create a monoalphabetic substitution cipher with variable-length cipher elements from English text. The cipher's length is 16,384 characters. We start by testing our model on the first 128 characters of the text, then we increase the length by a power of 2 until we reach 16,384. Figure \ref{fig:cipher-len} shows segmentation results for different cipher lengths. As expected, segmentation quality improves as cipher length increases. We notice the largest improvement going from 128 to 256 characters. Segmentation quality keeps improving until it almost plateaus after 2,048 characters.

\subsection{Monoalphabetic ciphers without word spaces}

We test our methods on the same set of 100 synthetic ciphers after removing spaces. To resemble real historical ciphers, we break the ciphertext into 43-character lines. The number of characters per line varies from one cipher to another, but as an approximation, we choose the average number of characters per line in a random sample of real ciphers. 

As shown in Tables~\ref{tab:mono-sp-f1} and~\ref{tab:mono-sp-seger} (second column), F1 and SegER scores for no-space monoalphabetic ciphers are generally slightly worse than ciphers with spaces. Our best performing model (Unigram~LM~2) achieves a SegER of 2.7\% on no-space monoalphabetic ciphers, which is very close to the 2.45\% on the same ciphers with spaces.

\subsection{Homophonic ciphers}

We test our segmentation methods on three real homophonic ciphers: S304, C13, and F283 (Table~\ref{tab:cipher-stats}). Note that S304 is the shortest and F283 is the longest of these ciphers (F283 is almost twice as long as S304).

As with previous experiments, we first evaluate how good the learned vocabularies are. Table~\ref{tab:seg-no-key-f1} shows F1 scores for different models. We notice that the 2-digit baseline is a strong baseline since most cipher elements are 2-digit in these historical ciphers. Our BPE and unigram models with the 2-digit heuristic give comparable results on ciphers S304 and C13, with an F1 score of 60-65\%. BPE~2 and Unigram~LM~2 give the highest F1 score on cipher F283. Overall, Unigram~LM~2 is the best performing model, with an average F1 score of 60\% over all three ciphers.

\begin{table}
	\centering
	\begin{tabular}{l|r|r|r}	
\hline
\multirow{2}{*}{\textbf{Model}} & \multicolumn{3}{c}{\textbf{Cipher Name}} \\ \cline{2-4}
 & \textbf{S304} & \textbf{C13} & \textbf{F283} \\ \hline
1-dig baseline  & 6.52 & 0.00 & 3.28 \\
2-dig baseline  & \textbf{63.95} & \textbf{64.95} & 41.94 \\
BPE  & 34.94  & 51.28  & 41.24 \\
BPE 2  & 61.45 & 64.62 & \textbf{53.61} \\
Unigram LM & 18.07 & 30.77 & 41.24 \\
Unigram LM 2 & 60.24 & 64.62 & \textbf{53.61} \\\hline
	\end{tabular}
	\caption{F1 \% ($\uparrow$) for segmenting three real homophonic ciphers using different models.}
	\label{tab:seg-no-key-f1}

\end{table}

We then evaluate the resulting segmentation for the three real ciphers. Table \ref{tab:seg-no-key-SegER} shows SegER scores for our models. The 2-digit baseline is much better than the 1-digit baseline on these historical ciphers, with an average improvement of more than 70\%. As we have seen in our synthetic, monoalphabetic cipher experiments, restricting piece length to a maximum of 2 improves performance for BPE and Unigram~LM. With the 2-digit heuristic, SegER improves by an average of 18\% and 58\% for BPE and Unigram~LM, respectively.

\begin{table}
	\centering
	\begin{tabular}{l|r|r|r}	
\hline
\multirow{2}{*}{\textbf{Model}} & \multicolumn{3}{c}{\textbf{Cipher Name}} \\ \cline{2-4}
 & \textbf{S304} & \textbf{C13} & \textbf{F283} \\ \hline
1-dig baseline & 164.15 & 204.91 & 213.14 \\
2-dig baseline & 60.00 & 41.11 & 64.19 \\
BPE & 78.07 & 51.80 & 50.10 \\
BPE 2 & 63.11 & 46.02 & 38.29 \\
Unigram LM & 84.59 & 72.85 & 38.19 \\
Unigram LM 2 & \textbf{46.67} & \textbf{20.83} & \textbf{14.95} \\\hline
	\end{tabular}
	\caption{SegER \% ($\downarrow$) for segmenting three real homophonic ciphers using different models.}
	\label{tab:seg-no-key-SegER}
\end{table}

While we could not find previously published work on this problem, we can see that our best method (Unigram~LM~2) achieves an average SegER of 27\% on the three real homophonic ciphers, with the best score of 14\% on the longest, 1,050-token F283 cipher.

\section{Related Work}

Previous decipherment work has mainly been focused on solving substitution ciphers with clearly segmented cipher elements, e.g.~\cite{hart-94, olsun-07, ravi-2008, corlett-2010, nuhn-2013, nuhn-2014, hauer-2014, aldarrab-2017}. Early decipherment approaches search for the substitution table that gives a highly probable plaintext according to a character LM. More recent approaches incorporate neural models. \citet{kambhatla-2018} use beam search and a neural LM to score candidate plaintext hypotheses from the search space for substitution ciphers. \citet{aldarrab-may-2021-sequence} view decipherment as a sequence-to-sequence translation problem. However, all of these works only deal with ciphers that have clear substitution units.

\citet{lasry-20} present an extensive study on papal ciphers from the 16th to 18th century. Those ciphers are numerical substitution ciphers that need to be segmented. For segmentation, \citet{lasry-20} create a set of segmenters (called ``parsers'' in the paper) from a collection of known cipher keys. Then they test the cipher in hand to see if any of the previously created segmenters can be a good fit. Our method, by contrast, is not limited to existing keys. In fact, our method is completely unsupervised and only uses ciphertext as input.

\section{Conclusion}

In this work, we present automatic methods for segmenting numerical substitution ciphers. We propose a method for solving non-deterministic substitution ciphers with existing keys using a lattice and a pretrained language model. Our method achieves a TER of 1.12\% on the IA cipher, a real historical cipher that has not been fully solved until this work.

We also propose a novel approach to segment numerical ciphers with no existing keys using subword segmentation algorithms. We use BPE and unigram language models as unsupervised methods to learn substitution units. We add a 2-digit heuristic based on historical cipher analysis. Our best method is able to segment 100 randomly generated monoalphabetic ciphers with an average SegER of less than 3\%, while still being robust to removing spaces. We test our methods on 3 real homophonic ciphers from the 16th-18th centuries. Our best method achieves an average SegER of 27\%, with a SegER of 14\% on the F283 cipher. To the best of our knowledge, this is the first work on automatically segmenting numerical substitution ciphers. 

\bibliography{acl_latex}
\end{document}